\documentclass[10pt,twoside, french]{article}

\usepackage{times}
\usepackage[utf8]{inputenc}
\usepackage[T1]{fontenc}
\usepackage{graphicx,amsmath}
\usepackage{floatrow}
\usepackage{comment}
\usepackage{latexsym}
\usepackage[table,xcdraw]{xcolor}

\newfloatcommand{capbtabbox}{table}[][\FBwidth]
\usepackage{multirow}
\usepackage{taln2019}
\usepackage[francais]{babel}

\newcommand{\x}{\mathbf{x}}
\newcommand{\wv}{\boldsymbol{\rho}} % word embedding, to be predicted 
\newcommand{\cwv}{\boldsymbol{\alpha}} % word embedding, for the context 

\title{Apprentissage de plongements de mots dynamiques avec régularisation de la dérive}
\author{Syrielle Montariol\up{1, 2}   \quad Alexandre Allauzen\up{1}\\
  {\small
    (1) LIMSI, CNRS, Univ. Paris-Sud, Univ. Paris-Saclay, F-91405 Orsay, France \\ 
    (2) Société Générale, 17 Cours Valmy 92043 Puteaux, France \\ 
    \texttt{
       syrielle.montariol@limsi.fr,  alexandre.allauzen@limsi.fr \\
}}}

\begin{document}
\maketitle

\resume{
    L'usage, le sens et la connotation des mots peuvent changer au cours du temps. Les plongements lexicaux diachroniques permettent de modéliser ces changements de manière non supervisée. Dans cet article nous étudions l'impact de plusieurs fonctions de coût sur l'apprentissage de plongements dynamiques, en comparant les comportements de variantes du modèle~\textit{Dynamic Bernoulli Embeddings}. Les plongements dynamiques sont estimés sur deux corpus couvrant les mêmes deux décennies, le \textit{New York Times Annotated Corpus} en anglais et une sélection d'articles du journal \textit{Le Monde} en français, ce qui nous permet de mettre en place un processus d'analyse bilingue de l'évolution de l'usage des mots.
}

\abstract{Learning dynamic word embeddings with drift regularisation}{
  Word usage, meaning and connotation change throughout time. Diachronic word embeddings are used to grasp these changes in an unsupervised way. In this paper, we use variants of the \textit{Dynamic Bernoulli Embeddings} model to learn dynamic word embeddings, in order to identify notable properties of the model. The comparison is made on the \textit{New York Times Annotated Corpus} in English and a set of articles from the French newspaper \textit{Le Monde} covering the same period. This allows us to define a pipeline to analyse the evolution of words use across two languages.
}

\motsClefs
  {Diachronie, Plongements lexicaux, analyse bilingue}
  {Diachrony, word embeddings, cross-lingual analysis}

\section{Introduction}

Les langues peuvent être considérées comme des systèmes dynamiques~: l'usage des mots évolue au cours du temps, reflétant les nombreux aspects des évolutions de la société, qu'ils soient culturels, technologiques ou dûs à d'autres facteurs \cite{Aitchison}.

La diachronie désigne l'étude de ces variations temporelles d'usage et de sens au sein d'une langue.
Ici, nous étudions un corpus journalistique d'une plage temporelle de deux décennies : l'usage des mots évolue suite à des événements ayant un retentissement médiatique.
Par exemple, l'usage du mot "Katrina" a connu un important changement au cours de ces deux décennies. Si par le passé, il fut exclusivement utilisé comme un prénom féminin, comme \textit{Justine} et \textit{Sonja}, dès  1999, son sens se rapproche de celui d'\textit{ouragan}, avec l'arrivée du premier orage tropical éponyme. Puis à partir de 2005 où le cyclone Katrina eut lieu, ce qui fut un prénom féminin partage désormais le même champs lexical que les mots "désastre", "dévastation" et  "inondation".

Détecter et comprendre ces changements avec le concours de méthodes d'apprentissage automatique est utile à la recherche linguistique, mais aussi à de nombreuses tâches de traitement automatique des langues. Ajouter une notion temporelle aux représentations de mots permet d'étudier des corpus qui s'étendent sur des plages temporelles longues avec une plus grande acuité. 
Le problème se pose particulièrement aujourd'hui, alors qu'un nombre croissant de documents historiques sont numérisés et rendus accessibles; leur analyse conjointe à celle de corpus contemporains, pour des tâches allant de la classification de documents à la recherche d'information, nécessite de prendre en compte la diachronie.

Suivant les travaux de~\shortcite{BengioNLM} puis~\shortcite{mikolovW2V}, de nombreuses méthodes  de représentations vectorielles de mots ont été mises au point depuis deux décennies. Elles permettent de représenter les mots par des vecteurs continus de faible dimension: les plongements lexicaux ou \textit{word embeddings}. Néanmoins, ces plongements lexicaux reposent sur l'hypothèse que le sens d'un mot est inchangé sur l'ensemble du corpus. Cette hypothèse d'une représentation statique peut s'avérer limitée. Ainsi en supposant qu'un changement dans le contexte usuel d'un mot reflète un changement dans la signification de ce mot, il est possible d'entraîner des plongements de mots diachroniques : qui évoluent au cours du temps en suivant les changements d'usage des mots.

Récemment \shortcite{rudolph2018dynamic} ont proposé un tel modèle, nommé \textit{Dynamic Bernoulli Embedding} (DBE). Il apprend des représentations de mots qui évoluent au cours du temps selon les strate temporelles d'un corpus, en caractérisant la dérive de ces représentation d'une strate à l'autre au moyen d'un processus aléatoire gaussien.
Des choix de modélisation différents ont été effectués par d'autres auteurs dans la littérature. Les approches de \shortcite{BBP-RujunHan} et \shortcite{Hamilton2016} impliquent d'apprendre des plongements lexicaux pour chaque strate temporelle sans les relier chronologiquement; tandis que \shortcite{kim2014} apprend les plongements diachroniques de façon incrémentale, mais sans contrôler la dérive de ces plongements.

Dans cet article nous prenons pour base le modèle DBE, qui présente un bon compromis entre simplicité et modulabilité, pour questionner l'importance de ces différents choix de modélisation. Dans ce but, nous analysons le comportement des plongements de mots appris à partir de ce modèle (décrit à la section~\ref{sec:dbe}) sur deux tailles de strates temporelles -- mensuelle et annuelle -- et l'appliquons (dans la section~\ref{sec:expe}) à des corpus dans deux langues différentes : français et anglais. Les données en anglais proviennent du \textit{New York Times Annotated Corpus} \footnote{https://catalog.ldc.upenn.edu/LDC2008T19}  \cite{nytLDC}, qui s'étend de 1987 à 2006; le corpus en français est constitué d'articles du journal \textit{LeMonde} collectés de façon à couvrir la même période. L'étude de ces deux corpus nous permet, dans un deuxième temps, d'étudier de façon conjointe l'évolution d'un mot à travers les deux langages.

%Dans cet article, la représentation vectorielle d'un mot est assimilée à son sens dominant au sein du corpus étudié. La détection d'une évolution dans l'un des sens d'un mot est donc simplifiée en la détection d'un changement dans ce sens dominant.

\section{État de l'art}
\label{sec:sota}

Les premières méthodes automatiques d'étude de la diachronie se basent sur la détection de changements dans les co-occurrences des mots, puis sur des approches basées sur la similarité distributionnelle \cite{Gulordava} en construisant des mesures d'information mutuelle à partir de matriças de co-occurrences.

L'usage de méthodes d'apprentissage automatique basées sur les plongements lexicaux est récent et a connu une forte hausse d'intérêt depuis deux ans, avec la publication consécutive de trois articles dédiés à l'état de l'art de ce domaine \cite{SOTAKutuzov, Tahmasebi2018SurveyOC, tang_2018}.

Dans un des premiers articles employant ce type de méthode~\cite{kim2014}, les auteurs estiment des plongements lexicaux pour la première strate temporelle $t_0$ puis mettent à jour ces plongements pour les strates temporelles suivantes, considérant les plongements au temps $t-1$ comme initialisation pour la strate $t$. D'autres travaux ont ensuite  vu le jour, reposant sur l'apprentissage de façon indépendante des plongements lexicaux pour chaque strate temporelle. Néanmoins, les plongements ainsi obtenus ne sont pas directement comparables car appartiennent à des espaces vectoriels différents. Deux approches sont alors envisageables : d'une part, déterminer la meilleure transformation linéaire afin d'aligner les espaces de représentation à travers les périodes~\cite{Hamilton2016,Dubossarsky,Szymanski, Kulkarni}; 
% Par exemple, \shortcite{Hamilton2016} utilisent le problème du Procuste orthogonal pour effectuer l'alignement. Des méthodes similaires sont utilisées par \shortcite{Dubossarsky,Szymanski, Kulkarni}. 
d'autre part, calculer la similarité cosinus entre chaque paire de mot à l'intérieur d'une strate temporelle, les similarités étant alors comparables d'une strate sur l'autre sans nécessiter d'alignement \cite{kim2014}.

Les méthodes dites dynamiques constituent un second type d'approche. Le corpus d'étude est toujours divisé en strates temporelles, mais cette fois les plongements lexicaux diachroniques sont appris de façon conjointe sur l'ensemble des strates. Ils sont ainsi placés dans un même espace de représentation dès l'apprentissage. Pour cela, \shortcite{bamler17a} utilisent des modèles bayésiens d'apprentissage de plongements lexicaux : les  vecteurs sont liées à travers les périodes à l'aide d'un processus de diffusion temporel qui contrôle leur évolution. Poursuivant le même objectif, différentes méthodes ont été proposées~\cite{yao2018dynamic, rudolph2018dynamic, BBP-RujunHan} afin de mettre en évidence de façon jointe l'évolution continue du sens des mots. Ces méthodes permettent de s'affranchir de la limite de volume de données par strate temporelle lors de l'apprentissage.
 
%Le \textit{New York Times Annotated Corpus} \cite{nytLDC} a été utilisé à plusieurs reprises dans la littérature pour entraîner des plongements de mots diachroniques. \shortcite{yao2018dynamic} appuient leur analyse sur un corpus du New York Times similaire à celui-ci, mais extrait à partir de l'API du journal sur une période plus longue (26 ans). \shortcite{Szymanski} et \shortcite{Zhang2015} s'appuient sur ce corpus pour résoudre une tâche d'analogies temporelles, et \shortcite{Azarbonyad} y observent les références au terrorisme et à l'islam avant et après les attentats du World Trade Center.

La majorité de ces modèles sont évalués sur des corpus en anglais. À notre connaissance, bien que plusieurs auteurs ont expérimenté sur d'autres langues que l'anglais ~\cite{Hamilton2016, eger-mehler-2016-linearity}, aucun travaux n'a tenté de comparer l'évolution de mots à travers plusieurs langues à l'aide de méthodes de plongements diachroniques.

% Mettre des références plus anciennes (avant 2001) et plus françaises. Par exemple, avec les travaux sur la datation des textes de la Bible au moyen de méthodes statistiques dès les années 70-80 à Lyon.

\section{Modèles de plongements de mots dynamiques}
\label{sec:dbe}
Nous partons du modèle \textit{Dynamic Bernoulli Embeddings} (DBE) de \shortcite{rudolph2018dynamic}. Il se base sur les plongements de mots de la famille exponentielle \cite{rudolph2016exponential}, qui sont une généralisation probabiliste du modèle \textit{Continuous Bag-of-Words} (CBOW) de \shortcite{mikolovW2V}. Nous en fournissons une brève description avant de présenter les différentes variantes mises en place.

\subsection{Le modèle DBE}

L'objectif de ce modèle est de prédire un mot à partir de son contexte. Afin de désigner un mot $v$ parmi un vocabulaire de taille $V$, on considère un vecteur de variables aléatoires binaires $\x_{v} \in \{0,1\}^V$, où seule la composante associée au mot $v$ vaut $1$. Le mot $v$ à la position $i$ dans le corpus est donc représenté par le vecteur binaire $\x_{iv}$ et son contexte  $\boldsymbol{c}_i$ est constitué des $C$ mots avant et des $C$ mots après ($C$ étant la taille de la fenêtre). Ainsi, $\x_{\boldsymbol{c}_i}$ regroupe l'ensemble des points constituant le contexte du mot $i$. 
Le modèle DBE prédit le vecteur binaire du mot $\x_{iv}$ à partir de son vecteur de contexte $\x_{\boldsymbol{c}_i}$ selon la loi de Bernoulli suivante: $\x_{iv} | \x_{\boldsymbol{c}_i} \sim Bern(p_{iv}) $. Le paramètre de la loi de Bernoulli  $p_{iv}$ est calculé à partir des plongements lexicaux $\wv_{v}$ du mot à prédire et $\cwv_{v'}$ des mots du contexte : 
\begin{equation}\label{eq:piv}
p_{iv} = \sigma
\Big( 
\wv_{v}^T ( \sum_{j \in \boldsymbol{c}_i} \sum_{v'\in V} \cwv_{v'} \x_{jv'})
\Big). 
\end{equation}

Ainsi la somme sur $v'$ sélectionne les plongements $\cwv_{v'}$ des mots du contexte, qui sont ensuite additionnés (somme sur $j$) afin de créer un vecteur représentant le contexte $\boldsymbol{c}_i$. Le paramètre de Bernoulli résulte de l'application de la fonction sigmoïde $\sigma$ au produit scalaire de ce vecteur avec le plongement $\wv_{v}$ du mot à prédire.

\textbf{Vers un modèle dynamique:}
Pour rendre ce modèle dynamique, considérons un corpus composé de $T$ strates temporelles indicées par $t$. Dans chaque strate, chaque mot $v$ a deux types de représentions : celle en tant que mot de contexte $\cwv_v$, et celle en tant que mot central $\wv_v$. Le vecteur $\cwv_v$ est considéré comme invariant : il est commun à toutes les strates temporelles. Seuls les plongements $\wv_v$ évoluent au cours du temps selon la marche aléatoire gaussienne suivante: 

\begin{equation}\label{eq:deriv}
    \wv_v^{(0)} \sim \mathcal{N}(0,\lambda_0^{-1}I), \textrm{ puis pour } \forall t \geq 1,\ \wv_v^{(t)} \sim \mathcal{N}(\wv_v^{(t-1)},\lambda^{-1}I).
\end{equation}
Le paramètre $\lambda$, nommé \textit{dérive}, est le même pour l'ensemble des strates et contrôle l'évolution du vecteur $\wv_v$ d'une strate temporelle sur l'autre.

\textbf{Apprentissage:}
L'apprentissage de ce modèle, plus précisément décrit par ~\shortcite{rudolph2018dynamic},  s'appuie sur une variante de la stratégie du \textit{negative sampling}~\cite{mikolovW2V}. L'objectif est d'optimiser la fonction  suivante: 
\begin{equation}
    \mathcal{L}(\wv,\cwv) =\mathcal{L}_{pos}(\wv,\cwv) + \mathcal{L}_{neg}(\wv,\cwv) +  \mathcal{L}_{prior}(\wv,\cwv)
\end{equation}
Le premier terme $\mathcal{L}_{pos}$ représente la log-probabilité associée aux exemples positifs, tandis que le second ($\mathcal{L}_{neg}$) correspond à celle associée à des exemples négatifs tirés aléatoirement. Le troisième terme agit comme un terme de régularisation sur $\cwv$ et sur la dynamique des plongements $\wv$, et consiste à pénaliser le vecteur $\wv_v ^{(t)}$ lorsqu'il s'éloigne trop fortement du vecteur $\wv_v ^{(t-1)}$, de la manière suivante: 
\begin{equation}\label{eq:prior}
    \mathcal{L}_{prior}(\wv,\cwv) = - \frac{\lambda_0}{2} \sum_v \| \cwv_v \|^2- \frac{\lambda_0}{2} \sum_v \| \wv_v ^{(0)} \|^2 - \frac{\lambda}{2} \sum_{v,t} \| \wv_v ^{(t)} - \wv_v ^{(t-1)} \|^2. 
\end{equation}

% Pour finir, à partir de la fonction objectif $\mathcal{L}(\rho,\alpha)$, les paramètres $\rho,\alpha$ sont mis à jours en remontant le gradient stochastique.

\subsection{Variantes de régularisation}

La première variante du modèle se rapproche du principe d'apprentissage incrémental proposé par \shortcite{kim2014}. Elle consiste à supprimer la régularisation sur la dérive des plongements de mots. Dans ce cas, la fonction de coût ne prend en compte que les deux premiers termes de la log-priore ainsi que  $\mathcal{L}_{pos}$ et  $\mathcal{L}_{neg}$. Par la suite, nous intitulons cette variante DBE-I (Incrémental).

La seconde variante consiste à abolir l'obligation de chronologie dans les vecteurs temporels successifs. En remplaçant la troisième composante de $\mathcal{L}_{prior}$ par $\sum_{v,t} \| \wv_v ^{(t)} - \wv_v ^{(0)} \|^2$, on force le vecteur $\wv_v ^{(t)}$ à rester proche du plongement d'origine $\wv_v ^{(0)}$. Ce principe est similaire à celui de \shortcite{BBP-RujunHan}, où les plongements de mots diachroniques sont appris de façon indépendante sur chaque strate temporelle. Cette variante est désignée par DBE-NC (Non Chronologique).

Une autre version de cette dernière fonction est mise en place, afin de prendre en compte l'éloignement temporel. Dans ce but, la troisième composante de la log-priore $\sum_{v,t} \| \wv_v ^{(t)} - \wv_v ^{(0)} \|^2$ est multipliée par un facteur temporel : le coefficient devient $ - \frac{\lambda }{2} * t$ et permet de contrôler l'éloignement à la priore sans ajouter de dépendance entre les strates temporelles successives. Cette dernière version est nommée DBE-SC (Semi Chronologique).

\section{Expérimentation}\label{sec:expe}

Nous expérimentons à partir de notre propre implémentation du modèle DBE en \textsc{pytorch}. Dans un premier temps, nous analysons de façon quantitative le comportement des différentes variantes du modèle DBE définies précédemment. Puis nous observons de plus près la dérive des mots; en particulier, nous mettons en place un processus pour comparer les évolutions des mots dans deux langues de façon conjointe.

\subsection{Données et hyper-paramètres}\label{ssec:data}

Les données du  \textit{New York Times Annotated Corpus} 
%people/allauzen/data/ldcxxx, Obtained on Limsi's user agreement with LDC.\\
sont composées de 1~855~000 articles s'étalant sur une période d'environs 20 ans, du $1^{er}$ janvier 1987 au 19 juin 2007. Le journal \textit{Le Monde} est un des quotidiens les plus lus en France; nous en collectons des articles entre le $1^{er}$ janvier 1987 et le 31 décembre 2006.
Ces deux corpus sont divisés en $T=20$ strates temporelles annuelles\footnote{La dernière année du corpus \textit{NYT} étant incomplète, elle n'est pas prise en compte dans l'analyse.} et $T=240$ strates temporelles mensuelles.

Pour construire le vocabulaire, nous sélectionnons pour les deux langues $V = 40~000$ mots selon leurs fréquences après avoir retiré les mots-outils. De même que \shortcite{mikolovW2V}, nous sous-échantillonnons les mots fréquents en retirant chaque mot $i$ avec une probabilité 
$p = 1 - \sqrt{\frac{10^{-5}}{\textrm{fréquence}(i)}}$. 
% fréquence minimale dans ce vocabulaire = 655 for nyt and 313 for leMonde
Dans le corpus \textit{LeMonde}, le nombre moyen de mots par strate temporelle est d'environ 3.5 millions pour les strates annuelles et 300k pour les strates mensuelles.
%3~526~246 (by year) and 293~853 (by month).
Dans le corpus \textit{NYT}, ce nombre est d'environ 9 millions pour les strates annuelles et 750k pour les strates mensuelles.
%8~998~071 (by year) and 749~839 (by month)
Le corpus est ensuite divisé en échantillons d'apprentissage, de validation et de test. Ces derniers comprennent chacune 10\,\% des données tirées aléatoirement. Les embeddings sont entraînés avec 1000 mini-batchs par strates temporelles pour l'analyse annuelle et 100 mini-batchs pour l'analyse mensuelle.

Les hyper-paramètres sont sélectionnés à partir de l'étude de la log-probabilité sur les exemples positifs $\mathcal{L}_{pos}$ calculée sur l'échantillon de validation de chaque corpus, à partir du modèle DBE classique. Afin de permettre la comparaison, les valeurs de $\mathcal{L}_{pos}$ sont mises à l'échelle selon la règle suivante :
$$\textrm{Échelle} = \frac{\textrm{Nb de mots dans l'échantillon de validation}}{\textrm{Nb de mots dans chaque mini-batch}}.$$

Dans un premier temps, le modèle est entraîné sur l'ensemble du corpus sans composante temporelle (modèle statique). Ainsi, les  plongements lexicaux  $\wv$ et $\cwv$ peuvent servir par la suite d'initialisation pour les modèles temporels. Suite à l'évaluation sur l'échantillon de validation, la fenêtre de contexte choisie est $C=4$\footnote{Deux mots précédant et deux mots suivant le mot central.}. La dimension des plongements lexicaux est de 100 et le nombre d'exemples négatifs tirés pour chaque exemple positif est fixé à 10. Pour finir, la dérive $\lambda =1$ est celle qui offre les meilleurs résultats. La dérive initiale $\lambda_0$ est fixée, comme le font \shortcite{rudolph2018dynamic}, à $\frac{\lambda}{1000}$.

%Résultats : (accuracy)
%En général, from\_prior True fait moins bien que false. C'est donc mieux de lier chronologiquement les time slices. Le contexte qui donne la meilleure accuracy, dans les deux cas, est cs = 4. La précision sig = 1 est la meilleure pour chrono (25.6). sig = 10 donne des résultats particulièrement mauvais (13.0 pour chrono, 18.5 pour nonchrono, étonamment meilleur. De même, pour sig = 5, résulats acceptables pour non chrono (23.8 en moyenne, encore mieux que sig = 1 où l'accuracy moy est de 23.1) mais très mauvais (15.4) pour chrono !
%From scratch (sans initialiser par le modèle statique) donne des résulats extrêmement mauvais aussi (accuracy = 11.8 en moyenne)

\subsection{Évaluation quantitative}

Dans un premier temps, nous analysons l'effet des différentes variantes de la fonction de coût sur les performances du modèle mesurées en terme de log-vraisemblance et sur la distribution des dérives des mots.

\subsubsection{Évolution de la log-vraisemblance}

\begin{figure}
\begin{center} 
\includegraphics[width=1\textwidth]{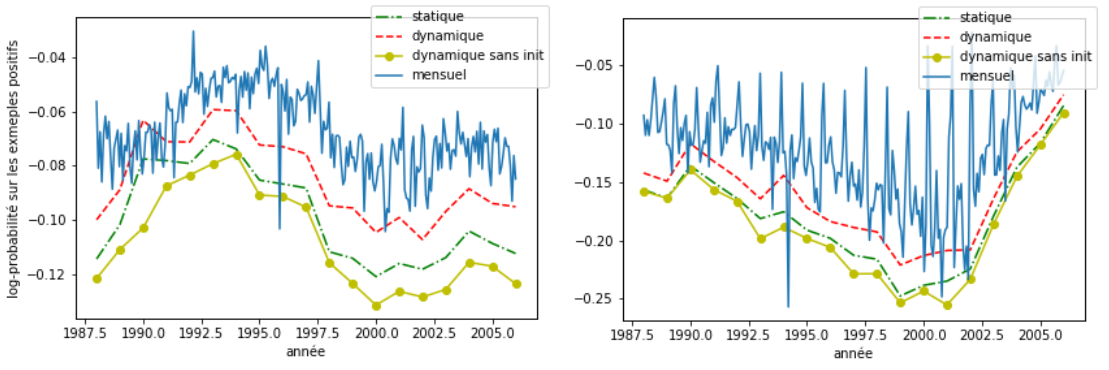}
\end{center} 
\caption{Log-probabilités sur les exemples positifs sur l'échantillon de test du corpus \textit{NYT} (à gauche) et \textit{LeMonde} (à droite), à partir des modèles statique, dynamiques, et dynamiques sans pré-entraînement pour des strates annuelles, et dynamique pour des strates mensuelles.}
\label{fig:likelihood}
\end{figure}

Dans cette partie, nous calculons la log-probabilité du modèle DBE sur les exemples positifs $\mathcal{L}_{pos}$ sur les données de test de chaque corpus, \textit{NYT} et \textit{LeMonde}, pour les deux tailles de strates temporelles (Figure \ref{fig:likelihood}). Nous appliquons le terme d'échelle décrit dans la partie \ref{ssec:data}.

Dans un premier temps, les plongements lexicaux sont appris sur l'ensemble du corpus de façon statique. Puis ils sont utilisés sur chaque strate annuelle du corpus pour calculer $\mathcal{L}_{pos}$. La courbe obtenue est plus basse que la courbe associée au modèle dynamique appris en initialisant les vecteurs à partir de ce modèle statique. On constate logiquement que l'apprentissage dynamique permet aux  plongements d'être adaptés à chaque strate temporelle, et donc plus efficace pour prédire les données de test. 

% Que dire sur la courbe mensuelle ?!
% The monthly likelihood follows the trend of the annual one.

À l'inverse, pour les deux corpus, le modèle dynamique sans initialisation a la performance la plus faible. Cette tendance est confirmée par la log-probabilité moyenne sur l'ensemble des strates temporelles (Table \ref{tab:loglik}). Une explication se trouve peut être dans le faible volume de données sur chaque strate. Quand à la performance du modèle statique, elle dépend de l'homogénéité temporelle du corpus étudié; nos deux corpus couvrent une plage de temps relativement faible, justifiant la performance du modèle statique par rapport à celle du modèle dynamique sans initialisation.

Comme le montre le tableau~\ref{tab:loglik}, les variantes du modèle définies par les différentes fonctions de coût ont des performances très proches; l'ajout du coefficient de dérive croissant (DBE-SC) au modèle non chronologique (DBE-NC) permet d'augmenter légèrement sa performance, mais dans l'ensemble, c'est le modèle sans régularisation sur la dérive (DBE-I) qui obtient le score le plus élevé quelle que soit la taille de la strate temporelle.

L'étude de la log-probabilité ne reflète qu'une vision globale des performances du modèle; afin de mieux comprendre son comportement, nous observons ensuite la distribution des dérives pour chaque variation du modèle.

\begin{table}[t]
\footnotesize
\centering
\begin{tabular}{l|c|c|c|c|}
\cline{2-5}
\multirow{2}{*}{}                                              & \multicolumn{2}{c|}{\textbf{NYT}} & \multicolumn{2}{c|}{\textbf{Le Monde}} \\ \cline{2-5} 
                                                               & annuel           & mensuel        & annuel             & mensuel           \\ \hline
\multicolumn{1}{|l|}{Statique}                                 & -0.09875         & -0.06771        & -0.1794            & -0.1259           \\ \hline
\multicolumn{1}{|l|}{DBE}                                & -0.08476        & \textbf{-0.06720}        & -0.1606            & -0.1231           \\ \hline
\multicolumn{1}{|l|}{DBE sans initialisation}                      & -0.10774        & -0.07305       & -0.1873            & -0.1498           \\ \hline
\multicolumn{1}{|l|}{DBE-I}                     &  \textbf{-0.08448}        & -0.06752       & \textbf{-0.1593}   & \textbf{-0.1227}  \\ \hline
\multicolumn{1}{|l|}{DBE-NC}                        &     -0.08517             &    -0.06817            & -0.1607            & -0.1236           \\ \hline
\multicolumn{1}{|l|}{DBE-SC} & -0.08455         & -0.06752       & -0.1598            & -0.1228           \\ \hline
\end{tabular}
\caption{Log-probabilités moyennes sur l'ensemble des strates temporelles, sur l'échantillon de test des deux corpus, pour les différentes variantes d'apprentissage et de fonction de coût du modèle DBE.}
\label{tab:loglik}
\end{table}

\subsubsection{Caractérisation de la dérive des plongmements}

\begin{figure}[t!]
\begin{center} 
\includegraphics[width=1\textwidth]{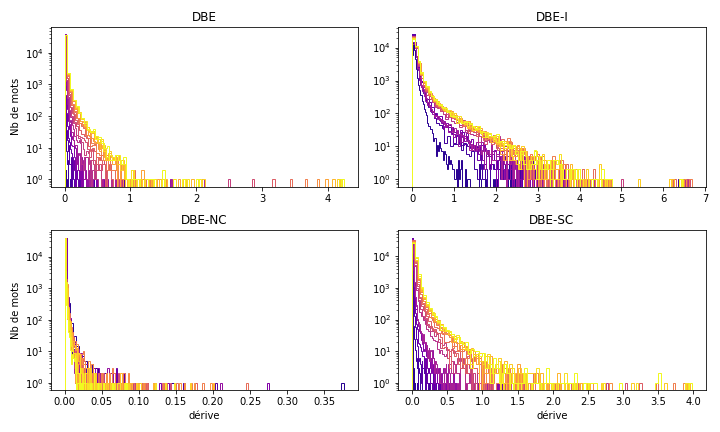}
\end{center} 
\caption{Histogramme des dérives entre les plongements de mots à $t_0=1987$  et à chaque strate temporelle annuelle successive du corpus \textit{LeMonde}, pour le modèles DBE et ses trois variantes. Plus la couleur est claire, plus la différence est calculée par rapport à une strate temporelle récente. Les nombres de mots (en ordonnée) sont en échelle logarithmique.}
\label{fig:freq}
\end{figure}

Dans le but d'analyser plus en détail le comportement du modèle et l'effet des variations de la fonction de coût, nous représentons les histogrammes superposés des dérives  successives observées sur le corpus \textit{LeMonde} (Figure \ref{fig:freq}). Les histogrammes pour le corpus \textit{NYT} présentent des tendances similaires, de même que le cas des strates temporelles mensuelles. La dérive de chaque mot est calculée  à partir de la distance euclidienne entre le plongement du mot au début du corpus $\wv_v^{(t_0)}$ et ses plongements successifs  $\wv_v^{(t)}$ à chaque nouvelle strate temporelle $t$. Sur les histogramme, les couleurs plus claires correspondent aux distributions des dérives aux strates récentes : ainsi, la courbe la plus claire représente la distribution des dérives de mots calculées entre $t_0 = 1987$ et $t = 2006$ tandis que la plus sombre représente la distribution des dérives entre $t_0 = 1987$ et $t = 1988$.

Une première propriété intéressante  est le caractère dirigé des dérives. Comme le montre le premier histogramme de la figure~\ref{fig:freq}, les valeurs des dérives augmentent à travers le temps pour le modèle DBE classique. Cela signifie que le modèle capture principalement des dérives possédant une tendance, plutôt que de brefs changements de plongements suivis de retours à la normale. 
Ainsi le terme de régularisation décrit par l'équation~\ref{eq:prior} réalise bien le compromis attendu, en considérant comme partie de l'objectif la détection des grandes tendances d'évolution du sens des mots et en omettant leurs brèves variations.

Ces brèves variations sont dues à des évènements qui modifient temporairement le contexte dans lequel apparaît un mot sans avoir un impact à long terme sur son sens. Elles sont plutôt capturées par la version DBE-NC du modèle, dont l'histogramme ne présente pas d'évolution dirigée de la dérive en fonction de la distance à $t_0$, donc ne distingue pas ces "bruits" de la tendance générale d'évolution des mots.
Pour finir, malgré l'absence de terme de régularisation sur la dérive, le modèle DBE-I capture naturellement une dérive relativement dirigée dans le temps bien que l'histogramme montre une plus grande sensibilité au bruit.

La seconde propriété mise en évidence est la capacité du modèle à distinguer les mots stables des mots dont l'usage évolue. En effet dans un intervalle de deux décennies, la majorité des mots est supposée peu évoluer. Le modèle DBE-NC, en introduisant une régularisation par rapport aux plongements de mots initiaux, permet de forcer le respect cette propriété : une grande part des mots sont presque invariants sur tout le corpus, et seule une sélection de dérives se démarque. Le modèle DBE classique permet aussi, dans une certaine mesure, de garder une faible dérive pour une grande partie des mots; de même pour le modèle DBE-SC. Seul le modèle DBE-I ne distingue pas naturellement les mots qui dérivent peu.

%Les deux histogrammes les plus à droite représentent les dérives des 2000 mots les moins fréquents du corpus (en haut) et les 200 mots les plus fréquents (en bas). Le modèle DBE a tendance à allouer faire dériver bien plus fortement les mots fréquents.
% The law of semantic change by Hamilton states the contrary. Here, is it just an effect of the corpus ?

\subsection{Évaluation qualitative}
\label{ssec:eval1}

La seconde étape de l'analyse est d'observer directement l'évolution des mots. 
À notre connaissance, il n'existe pas de corpus annotés permettant une évaluation directe des modèles diachroniques. Il est par contre possible d'observer l'évolution de certains mots choisis, permettant un premier diagnostic sous la forme d'une évaluation qualitative et subjective \cite{Tahmasebi2018SurveyOC}.
%Ce choix subjectif peut s'appuyer sur une intuition de changements passés pour des mots connus. 
Ainsi pour chaque variante du modèle, nous observons les mots qui dérivent le plus pour mieux en comprendre le comportement. Puis, afin d'étudier de façon conjointe l'évolution des mots sur les deux corpus, nous mettons en place un processus d'analyse diachronique inter-langues.

\subsubsection{Analyse des fortes dérives}

Nous nous concentrons ici sur le corpus \textit{LeMonde}. La période étudiée ne couvrant que deux décennies, on observe principalement des évolutions de contexte liées aux événements ayant un impact médiatique; les mots subissant de fortes dérives sont en majorité des entités nommées, et sont liés au contexte politique de la période, un thème récurrent dans ce journal.

Nous listons les 10 mots dont l'usage a le plus dérivé au cours des deux décennies selon chaque modèle dans le corpus, pour les deux tailles de strates temporelles. Du corpus \textit{NYT}, nous ne reportons que les 10 mots ayant le plus varié d'après le modèle DBE classique sur des strates temporelles annuelles (Table \ref{tab:top10drift}).

Dans le cas du modèle DBE, pour les strates mensuelles et annuelles, les mots qui dérivent le plus sont associés à des concepts ayant subi un changement notable et continu au cours de la période (\textit{euros}, \textit{al-qaïda}, \textit{internet}...). Cette observation est en accord avec la propriété observée précédemment au sujet du caractère dirigé des dérives. À l'inverse, les modèles DBE-I et DBE-NC annuels mettent principalement en valeur des dérives très fortes de mots liées à des événements uniques (\textit{zidane}, \textit{clearstream}, \textit{royal}). C'est particulièrement le cas pour le modèle DBE-NC sur les strates mensuelles , où les dérives rapportées sont sujettes à un bruit important.
% Cela vaudrait le coup, un jour, de regarder si ces mots ont bougé à un moment avec DBE classique puis sont resté stables. 

Afin de confirmer ces observations, le tableau~\ref{tab:norm-drift} moyenne pour chaque modèle les rapports entre la dérive moyenne (sur toutes les strates temporelles) et la dérive totale (entre la première et la dernière strate) des 10 et 500 mots qui évoluent le plus. Notons que les valeurs des dérives moyennes normalisées ne sont pas directement comparables, selon que l'on considère les strates annuelles ou mensuelles (car le nombre de strates diffère). La valeur moyenne pour les 10 mots qui dérivent le plus est toujours plus faible que celle sur les 500 mots qui dérivent le plus. Les dérives importantes sont donc les plus dirigées quel que soit le modèle. Le modèle DBE-I présente des valeurs très proches de celles du modèle DBE-SC, montrant que l'absence de régularisation par rapport à la dérive permet, dans une certaine mesure, de conserver une certaine robustesse au bruit, en adéquation avec les observations de la figure \ref{fig:freq}.

\begin{table}[!t]
\footnotesize
\centering
\begin{tabular}{|c||c|c|c||c|c|c|}
\hline
\textbf{NYT}                    & \multicolumn{6}{c|}{\textbf{Le Monde}}                                                                                                                                                                      \\[1ex] \hline
\textbf{Annuel}                 & \multicolumn{3}{c||}{\textbf{Annuel}}                                                                  & \multicolumn{3}{c|}{\textbf{Mensuel}}                                                               \\[1ex] \hline
\textbf{DBE}                    & \textbf{DBE}                     & \textbf{DBE-I}                   & \textbf{DBE-NC}                 & \textbf{DBE}                     & \textbf{DBE-I}                   & \textbf{DBE-NC}               \\ \hline
\cellcolor[HTML]{FFD87B}google  & \cellcolor[HTML]{FFD87B}euros    & clearstream                      & royal                           & \cellcolor[HTML]{FFD87B}euros    & \cellcolor[HTML]{FFD87B}rfa      & seznec                        \\ \hline
skilling                        & \cellcolor[HTML]{FFD87B}ump      & arcelor-mittal                   & \cellcolor[HTML]{FFD87B}sarkozy & \cellcolor[HTML]{FFD87B}sarkozy  & \cellcolor[HTML]{FFD87B}euros    & tramway                       \\ \hline
bloomberg                       & \cellcolor[HTML]{FFD87B}villepin & raimond                          & gdf                             & \cellcolor[HTML]{FFD87B}ump      & \cellcolor[HTML]{FFD87B}ségolène & pinochet                      \\ \hline
email                           & \cellcolor[HTML]{FFD87B}rfa      & shultz                           & \cellcolor[HTML]{FFD87B}euros   & francs                           & \cellcolor[HTML]{FFD87B}sarkozy  & hamas                         \\ \hline
\cellcolor[HTML]{FFD87B}katrina & \cellcolor[HTML]{FFD87B}sarkozy  & \cellcolor[HTML]{FFD87B}ségolène & hezbollah                       & \cellcolor[HTML]{FFD87B}ségolène & \cellcolor[HTML]{FFD87B}ump      & \cellcolor[HTML]{FFD87B}euros \\ \hline
cellphone                       & al-qaïda                         & outreau                          & liban                           & \cellcolor[HTML]{FFD87B}villepin & \cellcolor[HTML]{FFD87B}villepin & ahmadinejad                   \\ \hline
darfur                          & poutine                          & eads                             & thaksin                         & \cellcolor[HTML]{FFD87B}internet & monory                           & abbas                         \\ \hline
contras                         & gorbatchev                       & zapatero                         & \cellcolor[HTML]{FFD87B}ump     & ue                               & climatique                       & révision                      \\ \hline
blog                            & \cellcolor[HTML]{FFD87B}katrina  & \cellcolor[HTML]{FFD87B}villepin & islam                           & \cellcolor[HTML]{FFD87B}euro     & contras                          & fibre                         \\ \hline
\cellcolor[HTML]{FFD87B}euros   & \cellcolor[HTML]{FFD87B}internet & zidane                           & suez                            & bush                             & réévaluation                     & mahmoud                       \\ \hline
\end{tabular}
\caption{Listes des 10 mots ayant la plus grande dérive totale (distance entre la première et la dernière strate temporelle) pour les modèles DBE, DBE-I et DNE-NC sur le corpus \textit{LeMonde} et DBE sur le corpus \textit{NYT}. Les mots ayant un arrière-plan coloré sont communs à plus d'un modèle.}
\label{tab:top10drift}
\end{table}

%Détecter les mots qui dérivent ne suffit pas, il est important de comprendre le phénomène linguistique derrière. Un début d'analyse consiste à observer les 10 mots les plus similaires à chaque strates temporelle. On distingue alors entre quels champs lexicaux le mot a dérivé.
% un reviewer suggère d'aller voir les vecteurs de plus près. Récupérer la dimension qui a varié ? La lier avec les mots sur cette dimension ?

\begin{table}
\footnotesize
\centering
\begin{tabular}{l|c|c|c|c|}
\cline{2-5}
                                                               & \multicolumn{2}{c|}{\textbf{Annuel}} & \multicolumn{2}{c|}{\textbf{Mensuel}} \\ \cline{2-5} 
                                                               & \textbf{top 500}  & \textbf{top 10}  & \textbf{top 500}   & \textbf{top 10}  \\ \hline
\multicolumn{1}{|l|}{DBE}                                & 0.00642          & 0.00785          & 0.00356           & 6.280e-05        \\ \hline
\multicolumn{1}{|l|}{DBE-I}                     & 0.1152           & 0.0272           & 0.1149           & 0.0253         \\ \hline
\multicolumn{1}{|l|}{DBE-NC}                        & 0.5259           & 0.1054          & 0.0767           & 0.0530         \\ \hline
\multicolumn{1}{|l|}{DBE-SC} & 0.1096           & 0.0301          & 0.0715          & 0.0164          \\ \hline
\end{tabular}
\caption{Valeurs moyennes pour les 10 mots et les 500 mots dérivant le plus, de leur dérive moyenne normalisée. Les dérives sont calculées pour les 4 variantes du modèle DBE sur le corpus \textit{LeMonde}.}
\label{tab:norm-drift}
\end{table}

Pour finir, remarquons que parmi les mots qui ont le plus dérivé au cours des deux décennies, certains sont communs aux deux corpus (\textit{euros}, \textit{google}, \textit{katrina}). Nous proposons donc par la suite une méthode pour observer l'évolution conjointe de ces mots dans les deux langues.

\subsubsection{Analyse conjointe en anglais et français}

Dans cette partie, nous étudions l'évolution d'un mots en français dans le corpus \textit{LeMonde} et de sa traduction anglaise dans le corpus \textit{NYT}. Les modèles de plongements de mots étant appris de façon indépendante sur ces deux corpus, les vecteurs ne sont pas directement comparables. Nous effectuons alors un alignement des deux espaces de représentation en utilisant un dictionnaire bilingue comme outil de supervision \cite{muse}. Dans un premier temps, les plongements de mots des deux langues appris de façon statique sur l'ensemble du corpus sont normalisés; puis, l'espace de représentation des plongements en français est aligné sur l'espace vectoriel des plongements en anglais. Nous choisissons ce sens car les données du \textit{NYT} sur lesquelles les plongements anglais sont appris ont une volumétrie plus élevée, permettant des plongements lexicaux plus robustes. 

Nous utilisons l'outil \textsc{muse}\footnote{https://github.com/facebookresearch/MUSE} pour l'alignement. 
La supervision est effectuée au moyen d'un dictionnaire bilingue construit à partir des vocabulaires des deux corpus. Nous sélectionnons tous les mots ayant un équivalent dans l'autre langue à partir du dictionnaire fournit par \textsc{muse}, puis ajoutons manuellement une sélection de mots spécifiques aux données (principalement des entités nommées).
À partir des deux vocabulaires de 40~000 mots, nous obtenons un vocabulaire bilingue de 27~351 mots. Pour finir, les plongements sémantiques alignés $\wv_{\textrm{align}}$ and $\cwv_{\textrm{align}}$ sont utilisés pour initialiser les modèles dynamiques entrâinés sur chaque corpus.
%The held-out likelihood of the dynamic model initialised with the aligned embeddings is slightly lower than the one initialised without alignment. \\

Suite à l'apprentissage, pour chaque couple de mot dans le dictionnaire bilingue, nous calculons leurs dérives dans les deux corpus. Puis, nous calculons le cosinus comme une similarité entre les plongements des deux mots à la première et la dernière strate temporelle. Appelons cette valeur la similarité inter-langues. Nous calculons la dérive de cette similarité entre la première et la dernière strate temporelle, en mesurant la distance euclidienne entre ces deux valeurs. 

En observant la distribution de ces grandeurs, nous mettons en évidence 4 types de comportement de mot à travers les deux langues :
\begin{enumerate}
    \item Les mots qui dérivent dans la même direction dans les deux langues;
    \item Les mots qui dérivent dans les deux langages, mais dont le sens diverge (la similarité inter-langues décroît entre le première et la dernière strate temporelle);
    \item Les mots qui dérivent dans une seule des deux langues, tandis que l'autre reste stable;
    \item Les mots qui sont stables dans les deux langues.
\end{enumerate}

Nous différencions les classes de mots en utilisant la moyenne des dérive des mots de chaque langue ainsi que la moyenne de la dérive de la similarité inter-langues, et reportons leur répartition au sein du vocabulaire bilingue dans le tableau \ref{tab:cross-emb}. La majorité des mots appartiennent à la catégorie $(4)$ (mots stables dans les deux langues), ce qui confirme la propriété du modèle DBE énoncée à partir de la figure \ref{fig:freq}; tandis que les mots qui évoluent dans les deux langues (catégories $(1)$ et $(2)$) sont les plus rares.

\begin{table}[!ht]
\footnotesize
\centering
\begin{tabular}{|l|c|c|c|c|c|}
\hline
Classe & (1)        & (2)     & (3-fr) & (3-en)  & (4)  \\ \hline
Pourcentage  & 5.4        & 5.5     & 16.1   & 15.2     & 57.8 \\ \hline
Exemple    & renouvelable & soviétique & francs & patrie & savon \\ \hline
\end{tabular}
\caption{Proportion des mots dans les différentes classes de comportement de dérive inter-langues, avec un exemple pour chaque classe.}
\label{tab:cross-emb}
\end{table}

% pour rendre comparable les colonnes ont peut meltiplier par le nb de tme steps

\begin{figure}[!ht]
\begin{center} 
\includegraphics[width=0.6\textwidth]{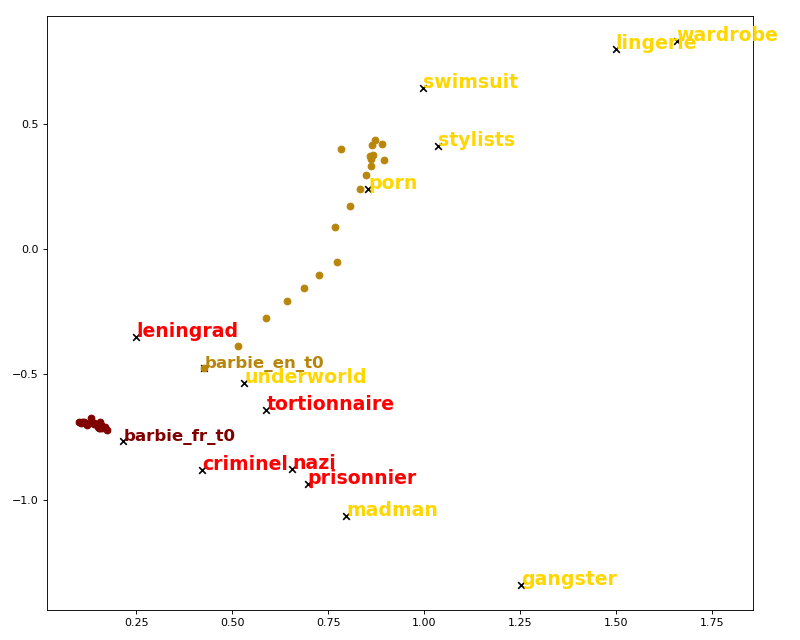}
\end{center} 
\caption{T-SNE des espaces de plongements sémantiques alignés, représentant l'évolution temporelle du mot \textit{Barbie} en français (rouge) et en anglais (jaune) ainsi que leurs plus proches voisins au cours du temps sur des strates annuelles.}
\label{fig:barbie}
\end{figure}

Considérons à titre d'exemple le mot \textit{Barbie}, qui appartient à la 3ème catégorie. 
L'espace aligné des plongements de mots  est réduit à deux dimensions au moyen de la méthode t-SNE ~\cite{tsne} pour représenter l'évolution de ce mot dans les deux langues (figure~\ref{fig:barbie}). Les mots les plus similaires au mot \textit{Barbie} dans chaque langue, et à chaque strate temporelle, sont indiqués sur le graphique. En français (en rouge), le mot cible ne subit pas d'évolution notable. Il est majoritairement associé au criminel nazi \textit{Klaus Barbie} et à son procès. Ces évènements ont eu une couverture médiatique moins importante et plus ponctuelle dans les journaux américains; l'équivalent anglais du mot cible évolue rapidement en direction du champs lexical de la mode, en association avec la célèbre marque de poupée homonyme. Son plongement sémantique se stabilise dans ce voisinage à partir des années 2000.

\section{Discussion}
Dans cet article, nous étudions en détail le comportement d'un modèle d'apprentissage de plongements lexicaux dynamiques. Le point de départ de notre étude est le modèle \textit{Dynamic Bernoulli Embeddings}, dont nous définissons plusieurs  variantes. Nous répliquons ainsi le comportement d'autres modèles d'apprentissage de plongements de mots diachroniques de la littérature. Deux propriétés nous paraissent importantes à distinguer pour bien caractériser ces modèles : la capacité à mettre en évidence des évolutions dirigées des plongements de mots, et la capacité à garder une partie du vocabulaire stable au cours du temps. 
Nous montrons ensuite qu'il est possible d'analyser l'évolution d'un mot dans deux langues de façon conjointe. Un processus d'analyse préliminaire est mis en place en initialisant les modèles dynamiques à partir de plongements de mots alignés, et en analysant la dérive de la similarité inter-langues.

Le domaine en plein essor qu'est l'apprentissage de plongements de mots dynamiques manque encore de la cohésion que possèdent les tâches plus anciennes du traitement automatique des langues. Les publications sur ce sujet portent sur des corpus très diversifiés et les évaluations se font le plus souvent de façon qualitative, en l'absence de base d'évaluation robuste. Notons qu'un cadre d'évaluation est difficile à définir dans le cas qui nous intéresse, tant les attentes applicatives vis-à-vis d'un modèle diachronique peuvent varier. De plus, un cadre mathématique commun et rigoureux n'a pas encore été défini \cite{SOTAKutuzov} et devrait s'appuyer sur les modèles d'apprentissage conjoint sur toutes les strates temporelles tel que celui décrit ici. 

En effet, l'apprentissage conjoint à travers toutes les strates permet de s'affranchir dans une certaine mesure de la nécessité d'avoir un grand volume de données dans chacune d'elle. Néanmoins, le caractère discontinu des strates temporelles induit le modèle à détecter seulement les dérives d'une strate à l'autre; plus les strates sont larges, plus les variations internes sont cachées, ou du moins moyennées. Ainsi, la question de la juste granularité temporelle se pose et dépend de l'application visée. Il est cependant important que les modèles étudiés puissent travailler à différents niveaux de finesse temporelle. Par exemple, lors de la recherche de dérives sémantiques brusques et de court terme, un type de modèle en temps continu~\cite{Rosenfeld2018DEEPNM} pourrait être plus adéquat, mais nécessite des informations temporelles très précises  qui en pratique se retrouvent presque exclusivement dans les corpus issus de médias sociaux. 

Une alternative est d'explorer l'usage de processus temporels de diffusion plus complexes, travaux initiés par exemple par~\cite{bamler17a} avec le processus d'Ornstein-Uhlenbeck. Enfin, l'emploi de strates temporelles non fixes dont les ruptures seraient apprises en même temps que les plongements lexicaux, assorti d'une régularisation sur la fonction de coût similaire à celle du modèle DBE-SC, serait une alternative à explorer. Néanmoins le cadre théorique approprié à cette sorte de quantification temporelle apprise par le modèle reste à définir.

% Un alternative sur un modèle appris conjointement sur toutes les time steps, serait d'utiliser des times tseps très courtes (si les metadatA le permettent) et détecter les strates où les plongements dérivent le plus, avant de réajuster les strates temporelles selon le schéma d'évolution détecté.

%La suite de ces travaux est d'amplifier notre démarche expérimentale en effectuant des évaluations avec d'autres jeux de données afin de rendre nos résultats comparables avec d'autres déjà publiés~\cite{Szymanski,Gulordava,yao2018dynamic}. 

%A future extension would be to use cross-lingual approaches to train dynamic and bilingual embeddings on both corpora jointly \cite{ruder2017survey}. Such method could better capture cross-lingual relations. However, a trade-off could be necessary to prevent the joint approach from softening the divergences of words across language while they drift. As a workaround, we plan to explore a  tailored regularisation scheme. 

%%================================================================
%\section*{Remerciements (pas de numéro)}

%Paragraphe facultatif, ajouté seulement dans la version finale (pas lors de la soumission).

%%================================================================
%% Note : si l'on préfère éviter de factoriser les crossrefs :
%% bibtex -min-crossrefs 99 taln-exemple
%%================================================================
\bibliographystyle{taln2019}
\bibliography{biblio}

\end{document}